\documentclass[letterpaper]{article} % DO NOT CHANGE THIS
\usepackage{aaai25}  % DO NOT CHANGE THIS
\usepackage{times}  % DO NOT CHANGE THIS
\usepackage{helvet}  % DO NOT CHANGE THIS
\usepackage{courier}  % DO NOT CHANGE THIS
\usepackage[hyphens]{url}  % DO NOT CHANGE THIS
\usepackage{graphicx} % DO NOT CHANGE THIS
\usepackage{natbib}  % DO NOT CHANGE THIS AND DO NOT ADD ANY OPTIONS TO IT
\usepackage{caption} % DO NOT CHANGE THIS AND DO NOT ADD ANY OPTIONS TO IT
\usepackage{algorithm}
\usepackage{newfloat}
\usepackage{listings}
\DeclareCaptionStyle{ruled}{labelfont=normalfont,labelsep=colon,strut=off} % DO NOT CHANGE THIS
\floatstyle{ruled}
\newfloat{listing}{tb}{lst}{}
\floatname{listing}{Listing}
\usepackage[noend]{algpseudocode}
\usepackage{amsfonts}
\usepackage{amsmath}
\usepackage{amssymb}
\usepackage{booktabs}
\usepackage[capitalize]{cleveref}
\usepackage{comment}
\usepackage{dsfont}
\usepackage{mathtools}
\usepackage[
    detect-all=true,
    detect-family=true,
    group-digits=false,
    separate-uncertainty=true,
    retain-zero-uncertainty=true
]{siunitx}
\usepackage{subcaption}

\usepackage{arydshln}
\usepackage{vector}

\DeclarePairedDelimiter{\norm}{\lVert}{\rVert}

\title{Enhanced Importance Sampling Through \\ Latent Space Exploration in Normalizing Flows}
\author{
    Liam Kruse, Alexandros Tzikas, Harrison Delecki, Mansur Arief, Mykel Kochenderfer
}
\affiliations{
    \textsuperscript{}Stanford University\\
    Department of Aeronautics and Astronautics\\
    Stanford, CA 94305 USA\\
    \{lkruse, alextzik, hdelecki, ariefm, mykel\}@stanford.edu
}

\begin{document}

\maketitle

%%%%%%%%%%%%%%%%%%%%%%%%%%%%%%%%%%%%%%%%%%%%%%%%%%%%%%%%%%%%%%%%%%%%%%%%%%%%%%%%
% Abstract
%%%%%%%%%%%%%%%%%%%%%%%%%%%%%%%%%%%%%%%%%%%%%%%%%%%%%%%%%%%%%%%%%%%%%%%%%%%%%%%%
\begin{abstract}
Importance sampling is a rare event simulation technique used in Monte Carlo simulations to bias the sampling distribution towards the rare event of interest.
By assigning appropriate weights to sampled points, importance sampling allows for more efficient estimation of rare events or tails of distributions. 
However, importance sampling can fail when the proposal distribution does not effectively cover the target distribution. 
In this work, we propose a method for more efficient sampling by updating the proposal distribution in the latent space of a normalizing flow. 
Normalizing flows learn an invertible mapping from a target distribution to a simpler latent distribution. 
The latent space can be more easily explored during the search for a proposal distribution, and samples from the proposal distribution are recovered in the space of the target distribution via the invertible mapping. 
We empirically validate our methodology on simulated robotics applications such as autonomous racing and aircraft ground collision avoidance.
\end{abstract}

%%%%%%%%%%%%%%%%%%%%%%%%%%%%%%%%%%%%%%%%%%%%%%%%%%%%%%%%%%%%%%%%%%%%%%%%%%%%%%%%
% Introduction
%%%%%%%%%%%%%%%%%%%%%%%%%%%%%%%%%%%%%%%%%%%%%%%%%%%%%%%%%%%%%%%%%%%%%%%%%%%%%%%%
\section{Introduction}
Safety-critical applications such as autonomous driving or aircraft controller design heavily rely on simulations to enhance safety through testing in controlled environments. 
Potential failures can be identified through simulation and then addressed before real-world deployment, reducing the risk of accidents \cite{corso2021survey}. 
Failures are often rare and safety thresholds are strict, so the events of interest---such as collisions or leaving a safe dynamic envelope---might be rarely encountered in simulation.
\textit{Importance sampling} (IS) is a variance-reduction technique used in Monte Carlo simulations to bias the sampling distribution towards the rare event of interest \cite{mcbook, corso2021survey}.
IS uses a \textit{proposal distribution} that focuses computational resources on scenarios likely to yield failure events, thus improving efficiency in failure detection. 
By assigning appropriate weights to sampled points, IS allows for more efficient estimation of the probability of failure compared to direct sampling from the target distribution.

\begin{figure}[ht]
    \centering
    \includegraphics[width=\columnwidth]{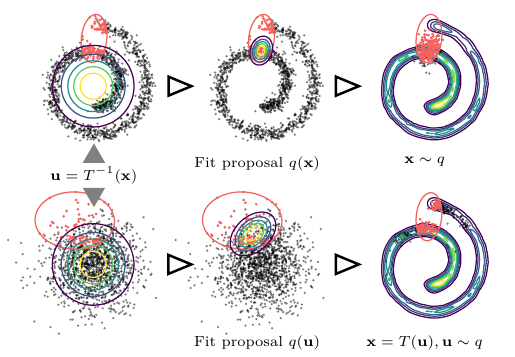}
    \caption{Importance sampling in target space (top row) versus importance sampling in a flow's latent space (bottom row). The target space proposal distribution generates many samples in the low-probability valley between the two failure modes, while the latent proposal generates samples that more closely map to the two failure regions.}
    \label{fig:overview}
\end{figure}

Importance sampling can fail when the proposal distribution does not effectively cover the target distribution. 
If the proposal distribution poorly matches the target distribution, the importance weights assigned to sampled points may become highly variable or imbalanced, leading to inaccurate failure estimates and increased estimate variance \cite{mcbook}. 
Furthermore, IS methods can miss less likely failure outcomes when the failure domain is multimodal, resulting in a biased estimate of the failure likelihood \cite{geyer2019cross}.

In this work we propose a technique to improve the efficiency of estimating the probability of failure by conducting importance sampling in the latent space of a normalizing flow. 
Normalizing flows are powerful generative models that learn an invertible mapping between a complex target density of interest and a simple, easy-to-evaluate latent density. 
We show that the distribution over simulation outcomes is mapped to a latent distribution that closely matches the IS prior, resulting in a balanced initial exploration over the space of possible outcomes. 
Moving forward, we refer to the space of simulation outcomes as the \textit{target space}.
We posit that the latent space of a normalizing flow is more effectively explored by IS methods than the target space.
Furthermore, we observe that latent space IS provides empirical advantages over target space IS, such as improved sample efficiency and better coverage of failure events, as shown in \cref{fig:overview}. 
Our specific contributions include the following:
\begin{itemize}
    \item We use the invertibility of normalizing flows to search for failure events in the latent space of flow models, demonstrating that importance sampling in latent space improves performance.
    \item We propose and justify an intuitive, easy-to-evaluate IS cost function formulation based on L{\"o}wner--John ellipsoids \cite{boyd2004convex} to facilitate latent space exploration.
    \item We evaluate our approach on simulated robotics applications including autonomous racing and aircraft ground collision avoidance.
\end{itemize}

%%%%%%%%%%%%%%%%%%%%%%%%%%%%%%%%%%%%%%%%%%%%%%%%%%%%%%%%%%%%%%%%%%%%%%%%%%%%%%%%
% Related Work
%%%%%%%%%%%%%%%%%%%%%%%%%%%%%%%%%%%%%%%%%%%%%%%%%%%%%%%%%%%%%%%%%%%%%%%%%%%%%%%%
\section{Related Work}
A rich body of literature exists on importance sampling methods, reflecting their widespread adoption for applications such as structural reliability analysis \cite{kurtz2013cross, papaioannou2016sequential, geyer2019cross} and safety validation for autonomous vehicles \cite{huang2019evaluation, corso2021survey}.  
IS methods can fail when the target distribution is not adequately explored, resulting in an \textit{inefficient} sampler, i.e., the effective sample size is small compared to the actual number of samples drawn \cite{corso2021survey}. 
Our proposed technique enhances IS coverage by fitting proposal distributions in the latent space of a normalizing flow, using the fact that the latent density is usually simpler than the target density.

Normalizing flows have been used to provide expressive proposal distributions in IS methods \cite{muller2019neural, gabrie2022adaptive, samsonov2022local}.
However, these approaches must interleave the sampling and training processes, resulting in complex training schemes and more strict assumptions to prove convergence \cite{samsonov2022local}.
Our proposed methodology involves importance sampling in the latent space of a \textit{pre-trained} flow model, providing more flexibility when the failure criteria or rare events of interest are updated yet the target distribution remains the same.
Furthermore, our approach is beneficial if the flow is costly to train from scratch.

Researchers have recently identified the utility of implementing Monte Carlo methods in the favorable geometry of normalizing flow latent spaces. 
\citet{hoffman2019neutra} propose a Hamiltonian Monte Carlo algorithm for sampling in the latent space of an inverse autoregressive flow \cite{kingma2016improved}, which can improve mixing speed. 
\citet{coeurdoux2023normalizing} propose a technique based on Langevin diffusion to correct for the topological mismatch between a latent unimodal distribution and a target distribution with disconnected support. 
Their goal is to limit out-of-distribution flow samples, whereas ours is to perform importance sampling for safety analyses.
\citet{noe2019boltzmann} perform Metropolis Monte Carlo in the latent space of a Boltzmann generator to generate independent samples of condensed matter systems and protein molecules. 
Meanwhile, \citet{choi2021featurized} map two densities to a shared latent feature space to obtain more accurate density ratio estimates.
Estimating density ratios is also a concern of \citet{sinha2020neural}, who perform bridge sampling after transforming the target space with a flow model.

Although importance sampling for rare events is the primary focus of this work, our methodology also lends itself to \textit{conditional flow sampling}, wherein the mapping between the latent and target spaces is conditioned on an input \cite{winkler2019learning}. 
Researchers have explored how the latent space can be partitioned to map different components of the input into disjoint regions in the target space \cite{dinh2019rad, winkler2019learning}. 
\citet{whang2021composing} perform approximate conditional inference for image inpainting by composing two flow models, while \citet{cannella2020projected} define a Markov chain within a flow's latent space to perform conditional image completion.
Our approach can be viewed as an approximate conditional sampling scheme, enabling the flow to more efficiently generate samples that satisfy user-defined requirements. %(e.g., a state variable falls within a particular range).

\section{Importance Sampling and Normalizing Flows}
We outline the fundamental theory behind importance sampling and normalizing flows before justifying the decision to perform IS in the latent space of a pre-trained flow model.

\subsection{Importance Sampling}
Simulations allow engineers to assess the performance of algorithms and models in diverse scenarios, including rare or dangerous scenarios that would be costly to replicate in real-world testing.
Assessing the probability of failure events can require a prohibitively large number of Monte Carlo simulations, especially if the event of interest is rare.

Consider an outcome space $\mathbf{x} \in \mathbb{R}^n$ with probability density function $p(\mathbf{x})$ and a \textit{cost function} $f(\mathbf{x})$ such that a \textit{failure event} occurs if and only if $f(\mathbf{x}) \leq 0$.
The probability of failure $P_F$ is given by the integral
\begin{equation}
\label{eq:pf-integral}
    P_F = \mathbb{E}_{p(\mathbf{x})} \left[ \mathds{1} \{ f(\mathbf{x}) \leq 0 \} \right] = \int \mathds{1} \{ f(\mathbf{x}) \leq 0 \} \cdot p(\mathbf{x}) d\mathbf{x}
\end{equation}
We can estimate $P_F$ via Monte Carlo simulations by drawing $N_s$ samples $\lbrace \mathbf{x}_1, \dots, \mathbf{x}_{N_s} \rbrace$ from $p(\mathbf{x})$ and taking the mean:
\begin{equation}
\label{eq:pf-mcs-estimate}
    \hat{P}_F = \dfrac{1}{N_s} \sum_{i=1}^{N_s}  \mathds{1} \{ f(\mathbf{x}_i) \leq 0 \}.
\end{equation}
This estimate is unbiased and has a coefficient of variation
\begin{equation}
    \delta_{\hat{P}_F} = \sqrt{\dfrac{1 - P_F}{N_s P_F}}.
\end{equation} 
Since the coefficient of variation is inversely proportional to the failure probability, many samples might be required to come up with a precise estimate of $P_F$, especially if $P_F$ is small \cite{papaioannou2016sequential}.

Importance sampling is a Monte Carlo simulation technique that aims to reduce the variance of $\hat{P}_F$ by sampling from an alternative sampling distribution---or \textit{proposal distribution}---denoted by $q(\mathbf{x})$. 
So long as the support of $q(\mathbf{x})$ contains the failure domain, the probability of failure integral in \cref{eq:pf-integral} can be rewritten as
\begin{equation}
    P_F = \int \dfrac{\mathds{1} \{ f(\mathbf{x}) \leq 0 \} \cdot p(\mathbf{x}) }{q(\mathbf{x})} \cdot q(\mathbf{x}) d\mathbf{x}.
\end{equation}
The importance sampling estimate of $P_F$ is given by
\begin{equation}
\label{eq:is-pf}
    \hat{P}_F = \dfrac{1}{N_s} \sum_{i=1}^{N_s} \mathds{1} \{ f(\mathbf{x}_i) \leq 0 \} \cdot \dfrac{p(\mathbf{x}_i)}{q(\mathbf{x}_i)}
\end{equation}
where the samples are distributed according to the proposal distribution $q(\mathbf{x})$. 
Thus, an appropriate choice of proposal distribution can reduce the variance of the estimate of $P_F$.

\subsection{Normalizing Flows}
Normalizing flows \cite{rezende2015variational} are a class of generative model used for density estimation and generative sampling. 
Normalizing flows transform a real vector $\mathbf{u}$ sampled from an easy-to-evaluate \textit{base distribution}, denoted by $p_\text{u}(\mathbf{u})$, through a transformation $\mathbf{z} = T(\mathbf{u})$ to produce a more expressive target density $p^*(\mathbf{z})$.
A common choice of base distribution is a standard normal distribution. 
The transformation must be \textit{invertible} and \textit{differentiable}, i.e., a \textit{diffeomorphism}. 
Imposing such topological constraints on a flow architecture ensures that the target density can be evaluated using the change of variables formula:
\begin{equation}
    \label{eq:change-of-var}
    p^*(\mathbf{z}) = p_\text{u}(\mathbf{u}) \lvert \det J_T(\mathbf{u})\rvert^{-1} \ \text{ where } \mathbf{z} = T(\mathbf{u}).
\end{equation}
The Jacobian of transformation $T$ is denoted by $J_T$; its determinant is a volume-correcting term that adjusts the probability density function of the transformed variable. 
The transformation (``flow'') itself is typically \textit{composed} of $D$ simpler transformations: $T = T_D \circ T_{D-1} \circ \dots \circ T_1$.
Since the transformations are composable and each step (``layer'') is a diffeomorphism, we can set $\mathbf{z}_0 = \mathbf{u}$, $\mathbf{z}_d = T_d \circ  \dots \circ T_1 (\mathbf{z}_0)$ and compute the Jacobian-determinant in the log domain as
\begin{equation}
    \label{eq:log-jac-det}
    \log \lvert \det J_T(\mathbf{u}) \rvert = \sum_{d=1}^D \log \lvert \det J_{T_d} (\mathbf{z}_{d-1}) \rvert.
\end{equation}

\begin{figure}[t]
    \centering
    \includegraphics[width=\columnwidth]{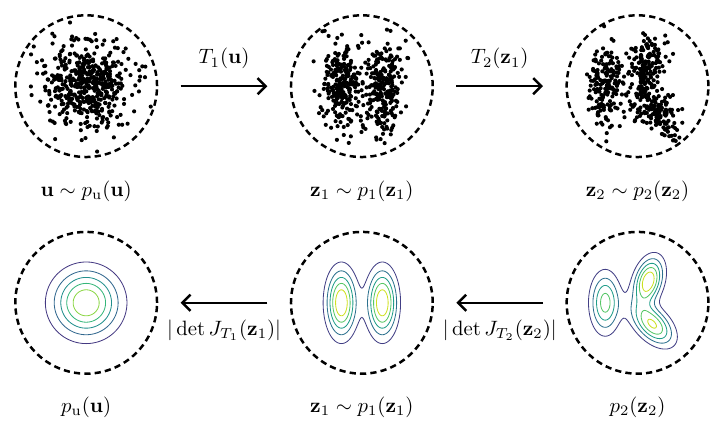}
    \caption{A normalizing flow transforms a base distribution to a target distribution.}
    \label{fig:normalizing-flow}
\end{figure}

\Cref{fig:normalizing-flow} shows the two operations provided by a normalizing flow. 
The \textit{forward} transformation is used when sampling, as an initial sample $\mathbf{u}$ is drawn from the base distribution and cascaded through the composed transformations. 
The \textit{inverse} transformation is used to perform density evaluation. 
The probability density function can be evaluated for an arbitrarily complex target distribution by iteratively computing the change of variables formula. 
Design considerations for both forward and inverse transformations are provided by \citet{papamakarios2021normalizing}.

\subsection{Justification of Latent Space Sampling}
Since normalizing flows learn an invertible mapping, we can fit proposal distributions over either the target space or the learned latent representation. 
As shown in \cref{fig:overview}, the target distribution can be highly non-isotropic in nature, with regions of extremely low density bordered by regions of very high likelihood. 
Thus, naively fitting a parametric proposal density can result in many generated samples that fall in regions of near-zero likelihood. 
Meanwhile, the latent distribution of normalizing flows is typically a standard normal, which is also a common choice for importance sampling priors \cite{papaioannou2016sequential}. 
Probability mass in a small neighborhood around a latent data point is nearly isotropic, which facilitates exploration by isotropic proposals \cite{hoffman2019neutra, cannella2020projected}. 

%%%%%%%%%%%%%%%%%%%%%%%%%%%%%%%%%%%%%%%%%%%%%%%%%%%%%%%%%%%%%%%%%%%%%%%%%%%%%%%%
% Methodology
%%%%%%%%%%%%%%%%%%%%%%%%%%%%%%%%%%%%%%%%%%%%%%%%%%%%%%%%%%%%%%%%%%%%%%%%%%%%%%%%
\section{Methodology}
In this section, we formalize our proposed technique for enhanced importance sampling and present our flow architecture, cost function formulation, and IS proposal-fitting methods.

\subsection{Normalizing Flow Architecture}

Recall that the flow transformation must be both invertible and differentiable.
A sufficient condition for invertibility is enforcing the transformation to be monotonic. 
\citet{durkan2019neural} propose the use of monotonic rational quadratic splines as building blocks for the transform.
In this work we use piecewise rational quadratic \textit{coupling} transforms to compose the flow models. 
Coupling transforms operate by splitting the input into two parts and applying an invertible function to one part while leaving the other part unchanged, thus ensuring efficient inversion and sampling capabilities \cite{dinh2016density, papamakarios2021normalizing}. 
The flow parameters are optimized by minimizing the forward KL divergence between the learned and target distributions.

\subsection{Cost Function Formulation}
The cost function should be continuous and bias the search over simulator outcomes towards the rare or failure events of interest \cite{corso2021survey}. 
However, computing the distance function for an arbitrary set is a hard problem, and even determining if an outcome falls inside or outside a set can be computationally expensive \cite{hormann2001point}. 
We propose a cost function geometry based on L{\"o}wner--John ellipsoids, i.e., the minimum volume ellipsoid that contains a set of points $\mathcal{X}$ \cite{boyd2004convex}. 
We posit that minimum volume ellipsoids are a reasonable choice of cost function since 1) they overapproximate the convex hull of $\mathcal{X}$, which is desirable for enforcing conservative safety thresholds, and 2) Gaussian mixture models are commonly used for proposal distributions, and each mixture component has ellipsoidal level sets.

Consider a finite set of points $\mathcal{X} = \{\mathbf{x}_1, \ldots, \mathbf{x}_m\} \subseteq \mathbb{R}^n$. The problem of finding the minimum volume ellipsoid that covers $\mathcal{X}$ is a convex optimization problem \cite{boyd2004convex}:
\begin{equation}
\label{eq:mve}
    \begin{array}{ll}
    \mbox{minimize}   & \log \det \vect{A}^{-1} \\
    \mbox{subject to} & \norm{\vect{A} \mathbf{x}_i + \vect{b}}_2 \leq 1, \quad i = 1,\ldots,m
    \end{array}
\end{equation}
where the variables are a symmetric matrix $\vect{A} \in \mathbb{R}^{n \times n}$ and $\vect{b} \in \mathbb{R}^n$. This formulation is attractive because the L{\"o}wner--John ellipsoid can be found quickly using standard optimization solvers such as \texttt{CVXPY} \cite{diamond2016cvxpy}.

A question that naturally arises is how to obtain a set of points $\mathcal{X}$ that is representative of the failure region. 
If the data used to train the flow model is available, then one solution is to construct $\mathcal{X}$ from the failures in this dataset.
However, if the dataset is relatively small or the failure events are extremely rare, then the set $\mathcal{X}$ might not adequately cover the entire failure domain. 
In this event, the overapproximating nature of L{\"o}wner--John ellipsoids is advantageous.
Another solution is to specify the set $\mathcal{X}$ manually based on user insight (e.g., defining the corners of a safety threshold hypercube).

Once $\mathcal{X}$ is obtained, the representative points are mapped deterministically to a set $\mathcal{U}$ in latent space via the invertible flow mapping. 
Solving \cref{eq:mve} with input $\mathcal{U}$ yields a L{\"o}wner--John ellipsoid in latent space. 
The Mahalanobis distance $d_M$ between the ellipsoidal region and a sample from the IS proposal density is easily computed.
If $d_M \leq 0$, the sample lies within the ellipsoid and is classified as a failure event.

\subsection{Importance Sampling Methods}
In this work we evaluate two importance sampling methods; however, any IS algorithm could be used in practice \cite{mcbook, corso2021survey}.
The first algorithm is the \textit{cross entropy} (CE) method \cite{de2005tutorial, geyer2019cross}, which attempts to learn the parameters of a parametric proposal distribution that minimizes the KL divergence between the optimal IS density and the proposal.
The CE method introduces a series of intermediate failure domains that gradually approach the true failure domain. 
At each step, the intermediate failure region is defined such that $\rho \cdot N_s$ samples fall in the region, where the $\rho$-quantile is chosen by the user.
The proposal distribution parameters are then fit via maximum likelihood estimation over these samples. 
The expectation-maximization algorithm is often used to fit the search distribution, though it must be adjusted to account for importance-weighted samples \cite{geyer2019cross}.

The second method is \textit{sequential importance sampling} (SIS) \cite{del2006sequential, papaioannou2016sequential}. 
Like the CE method, SIS introduces a series of intermediate failure distributions that gradually approach the optimal IS density. 
Samples for each intermediate distribution are obtained by resampling weighted particles from the previous distribution and then moving the particles to regions of high likelihood under the next failure distribution via Markov chain Monte Carlo. 
In this work, we use a conditional sampling Metropolis--Hastings algorithm to move the samples \cite{papaioannou2016sequential}. 

\subsection{Algorithm}
\Cref{alg:latent-is} presents the proposed latent IS methodology for estimating $P_F$. 
The algorithm takes as input a flow model and an importance sampling algorithm and outputs an estimate of $P_F$.
Furthermore, failure events can be easily generated after learning the proposal distribution by generating samples in latent space and mapping them back to target space. 
Without loss of generality, the user can input a set of failure sets $\{\mathcal{X}_i\}_{i=0}^n$, $\mathcal{X}_i = \{ \mathbf{x}^i_1, \ldots, \mathbf{x}^i_m\}$, with each set corresponding to a failure mode.
The L{\"o}wner--John ellipsoid is solved for each failure set, and the cost function computes the Mahalanobis distance for each ellipsoid and returns the minimum value.

\begin{algorithm}[t]
\caption{Latent Space Importance Sampling}
\label{alg:latent-is}
\begin{algorithmic}[1]
\Require{flow $T$, failure set $\mathcal{X} = \{\mathbf{x}_1, \ldots, \mathbf{x}_m \}$, proposal $q_0(\mathbf{u}; \vect{\theta}_0)$, latent flow distribution $p(\mathbf{u}) = \mathcal{N}(\mathbf{u}; \vect{0}, \mathbf{I})$}
\Ensure{probability of failure estimate $\hat{P}_F$}
\Procedure {Latent Importance Sampling}{}
\State $\mathcal{U} \leftarrow T^{-1}(\mathcal{X})$
\State $\vect{A}, \vect{b} \leftarrow $ solve for L{\"o}wner--John ellipsoid of $\mathcal{U}$
\State $\vect{\Sigma} = \left( \vect{A}^\top \vect{A} \right)^{-1},~\vect{\mu} = -\left( \vect{\Sigma} \vect{A}^\top \right) \vect{b}$
\State define function $f(\mathbf{u}) = \sqrt{(\mathbf{u} - \vect{\mu}) \vect{\Sigma}^{-1} (\mathbf{u} - \vect{\mu})^\top}$
\For {$k \leftarrow 1 : k_{\mathrm{max}}$}
\State $\text{elite samples $\mathbf{u}_e$}, ~\text{importance weights $\vect{w}$} \leftarrow$ \State\hspace\algorithmicindent $\textsc{ISMethod}(f(\mathbf{u}), q_k(\mathbf{u}; \vect{\theta}_k), p(\mathbf{u}))$
\While{\text{not converged}}
\State $\vect{\theta}_{k+1} \leftarrow \text{fit proposal parameters with } \mathbf{u}_e, \vect{w}$
\EndWhile
\EndFor
\State compute $\hat{P}_F$ with \cref{eq:is-pf}
\EndProcedure
\end{algorithmic}
\end{algorithm}
\vspace{-1em}

%%%%%%%%%%%%%%%%%%%%%%%%%%%%%%%%%%%%%%%%%%%%%%%%%%%%%%%%%%%%%%%%%%%%%%%%%%%%%%%%
% Experiments
%%%%%%%%%%%%%%%%%%%%%%%%%%%%%%%%%%%%%%%%%%%%%%%%%%%%%%%%%%%%%%%%%%%%%%%%%%%%%%%%
\section{Experiments}
This section presents our simulated robotics datasets and evaluation metrics before discussing experimental results.

\subsection{Data Simulators}
We use three autonomous system simulators to validate our proposed approach. 

\subsubsection{Nonholonomic Robot}
Consider a nonholonomic robot that moves in two dimensions. 
The three-dimensional robot state $s$ is defined as
\begin{align*}
    s &= \begin{bmatrix} x & y & \theta \end{bmatrix}^\top,
\end{align*} 
where $x$ and $y$ are the x- and y-positions of the vehicle and $\theta$ is the heading angle.
The control vector $a$ is
\begin{align*}
    a &= \begin{bmatrix} v & \alpha \end{bmatrix}^\top 
\end{align*}
with velocity $v$ and angular rate $\alpha$ in radians per second. 
We simulate noisy dynamics over a $T = 40$ second horizon, holding the inputs at constant magnitude for the duration of each simulation. 
However, to induce a multimodal outcome, we stochastically flip the sign of the angular rate input 15 seconds into each trial.
We train a flow on \num{e5} trials and save \num{e6} simulations for Monte Carlo evaluations.

\subsubsection{Cornering Racecar}
We next simulate data from a nonlinear single-track racecar modeled with a variant of the Fiala brush tire model \cite{subosits2021impacts}. 
The car uses model predictive path integral control \cite{williams2018information} to drift around a corner without spinning out of control. 
The eight-dimensional car state $s$ is defined as
\begin{align*}
    s &= \begin{bmatrix} x & y & \psi & v_x & v_y & \dot{\psi} & \delta & F_p \end{bmatrix}^\top,
\end{align*} 
where $x$ and $y$ are the x- and y-positions, $v_x$ and $v_y$ are the longitudinal and lateral velocities, $\psi$ and $\dot{\psi}$ are the yaw and yaw rate, $\delta$ is the steering angle, and $F_p$ is an input from the pedals.
We use the experimental setup of \citet{asmar2023model}, who provide an open-source Julia repository for their experiments.\footnote{\url{https://github.com/sisl/MPOPIS}}
The flow is trained on \num{5e5} trials, with \num{1.05e6} simulations saved for Monte Carlo evaluations.

\subsubsection{F-16 Ground Collision Avoidance}
Finally, we simulate an F-16 figher jet controlled by a ground collision avoidance system (GCAS). 
We use the dynamics model introduced by \citet{heidlauf2018verification} and the \texttt{JAX} code implementation\footnote{\url{https://github.com/MIT-REALM/jax-f16}} by \citet{so2023solving}. 
The F-16 begins in a dive towards the ground; the GCAS system then rolls the aircraft until the wings are level and pulls the nose above the horizon. 
We add noise to the F-16's initial state, thus simulating an envelope of possible start positions from which the GCAS must execute a recovery. 
The post-processed state vector is 12-dimensional, including the aircraft's roll, pitch, yaw, altitude, and airspeed. 
For a full list and definition of state variables, please refer to \citet{heidlauf2018verification}. 
Note that we remove the engine power lag variable (which is discrete), the stability roll rate and side acceleration/yaw rate integrators (which are observed to be close to zero), and the angle of attack variable. 
We train on \num{e6} trials and save \num{e7} simulations for Monte Carlo evaluations.

\subsection{Metrics}
We compute a reference failure probability for each dataset using the Monte Carlo estimate given in \cref{eq:pf-mcs-estimate} and then calculate the importance sampling estimate $\hat{P}_F$ using \cref{eq:is-pf}.
We return the relative error between these two values; a positive value indicates that the estimated failure likelihood is an overestimate of the true value. 
Next, we compute a series of metrics to evaluate the quality of the learned proposal density:
\begin{itemize}
    \item \textit{Average negative log likelihood (NLL)}: An evaluation batch of size $N_\text{eval}$ is drawn from the learned proposal distribution and the negative log-likelihood of each data point is computed according to \cref{eq:change-of-var} and \cref{eq:log-jac-det}. A \textit{lower} value indicates that the samples more closely match the learned flow density.
    \item \textit{Coverage}: The coverage metric proposed by \citet{naeem2020reliable} measures the fraction of real samples whose neighborhood contains at least one generated sample; it is useful for detecting mode dropping. A \textit{higher} value indicates better coverage.
    \item \textit{Density}: The density metric proposed by \citet{naeem2020reliable} rewards the proposal distribution for generating samples in regions where real data points (i.e., true failure events) are closely packed. A \textit{higher} value indicates better performance.
\end{itemize}

Lastly, we report the average number of samples until convergence, $\bar{N}_\text{total}$, to evaluate sample efficiency. 
A \textit{lower} value indicates that the method requires fewer samples and function evaluations to find a sufficient proposal distribution.

\subsection{Experimental Setup}
\begin{table*}[t]
    \centering
    \sisetup{round-mode=places, round-precision=4, table-align-uncertainty=true, separate-uncertainty=true}
    \begin{tabular*}{\textwidth}{@{\extracolsep{\fill}}
            l
            S[table-format=2.3(4), separate-uncertainty=true, retain-zero-uncertainty=true]
            S[table-format=3.3(5), separate-uncertainty=true, retain-zero-uncertainty=true]
            S[table-format=1.3(4), separate-uncertainty=true, retain-zero-uncertainty = true]
            S[table-format=1.3(4), separate-uncertainty=true, retain-zero-uncertainty = true]
            r
        @{}}
        \toprule
        {Method} & {$(\hat{P}_F - P_F)/P_F$} & {avg NLL$(\hat{x})$} & {coverage} & {density} & {$\bar{N}_\text{total}$} \\
        \midrule
        \multicolumn{6}{c}{{\textit{Nonholonomic Robot (3D)}: $P_F = 0.0072$}} \\
        {Latent-CE} & 0.480 \pm 0.075   & 3.432 \pm 0.259   & \bfseries 0.791 \pm 0.049 & 0.984 \pm 0.042   & \bfseries 4560 \\
        {Latent-SIS} & 0.479 \pm 0.118  & \bfseries 3.098 \pm 0.131 & 0.707 \pm 0.031   & \bfseries 0.990 \pm 0.010 & 9330 \\
        \hdashline
        {Target-CE} & -0.539 \pm 0.272  & 3.441 \pm 0.822   & 0.378 \pm 0.222   & 0.972 \pm 0.091   & 11040 \\
        {Target-SIS} & \bfseries 0.058 \pm 0.147 & 7.969 \pm 0.507  & 0.565 \pm 0.043   & 0.584 \pm 0.065   & 12030 \\

        \midrule
        \multicolumn{6}{c}{{\textit{Cornering Racecar (8D)}: $P_F = 0.0062$}} \\
        {Latent-CE} & -0.040 \pm 0.011  & \bfseries 8.435 \pm 0.088 & \bfseries 0.813 \pm 0.010 & 1.058 \pm 0.015   & \bfseries 29900 \\
        {Latent-SIS} & -0.036 \pm 0.032 & 8.711 \pm 0.148   & 0.791 \pm 0.012   & \bfseries 1.080 \pm 0.021 & 61200 \\
        \hdashline
        {Target-CE} & \bfseries -0.010 \pm 0.661 & 12.333 \pm 1.445 & 0.555 \pm 0.139   & 1.009 \pm 0.128   & 34200 \\
        {Target-SIS} & 0.060 \pm 0.280  & 12.835 \pm 1.025  & 0.706 \pm 0.032   & 0.886 \pm 0.024   & 63150 \\

        \midrule
        \multicolumn{6}{c}{{\textit{F-16 Ground Collision Avoidance (12D)}: $P_F = 0.0043$}} \\
        {Latent-CE} & -0.615 \pm 0.001  & \bfseries -32.455 \pm 0.068 & \bfseries 0.586 \pm 0.008 & 0.668 \pm 0.008 & \bfseries 300000 \\
        {Latent-SIS} & \bfseries -0.614 \pm 0.003 & -32.107 \pm 0.116   & 0.575 \pm 0.006   & \bfseries 0.669 \pm 0.006 & 622500 \\
        \hdashline
        {Target-CE} & -1.000 \pm 0.000  & 22.893 \pm 12.443 & 0.023 \pm 0.016   & 0.121 \pm 0.183   & 895000 \\
        {Target-SIS} & -0.999 \pm 0.003 & 37.255 \pm 0.333  & 0.095 \pm 0.011   & 0.028 \pm 0.004   & 652500 \\
        \bottomrule
    \end{tabular*}
    \caption{Experimental Results}
    \label{tab:metrics}
\end{table*}

We perform cross-entropy importance sampling and sequential importance sampling on all three datasets; furthermore, we evaluate each method in both latent space and target space.
For the target space experiments, we compute the L{\"o}wner--John ellipsoids on $\{\mathcal{X}_i,~i=0, \ldots, n\}$ directly.
In the latent experiments, we first compute $\mathcal{U} = T^{-1}(\mathcal{X})$ and then solve \cref{eq:mve} with constraints based on the points in $\mathcal{U}$.

\begin{figure}[H]
    \begin{subfigure}[t]{0.49\columnwidth}
        \centering
        \includegraphics[width = 0.99\textwidth]{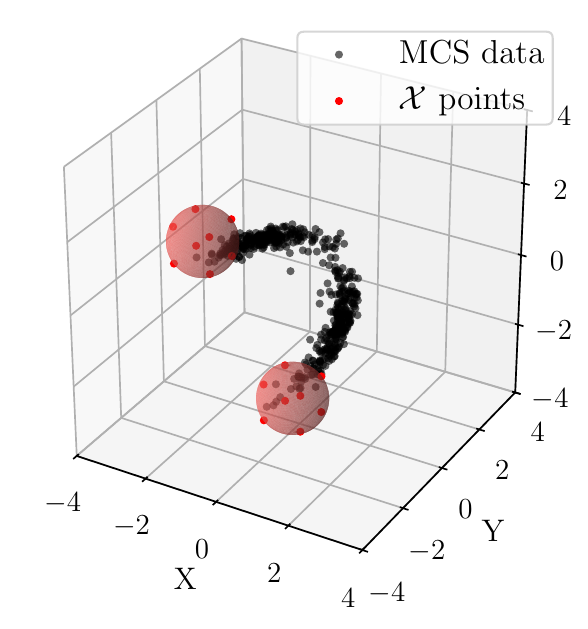}
        \caption{Target Space}
        \label{fig:target-roi}
    \end{subfigure}
    \hfill
    \begin{subfigure}[t]{0.49\columnwidth}
        \centering
        \includegraphics[width = 0.99\textwidth]{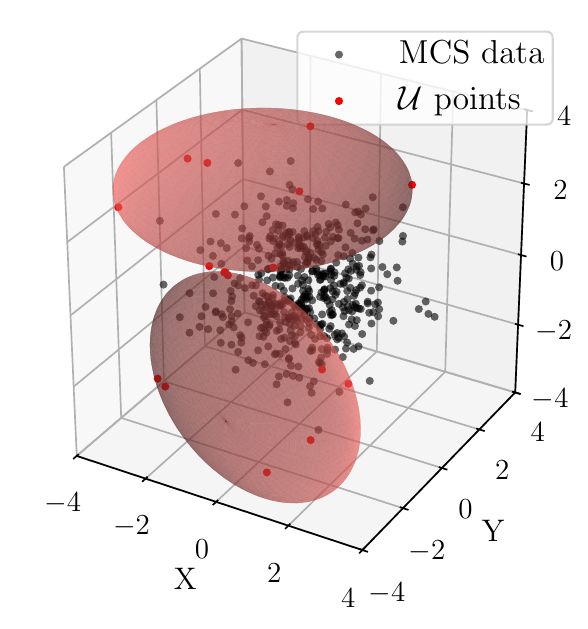}
        \caption{Latent Space}
        \label{fig:latent-roi}
    \end{subfigure}
    \caption{Failure regions for the nonholonomic robot shown in target space and latent space. The points in $\mathcal{X}_i$ are mapped to latent space and the L{\"o}wner--John ellipsoids are re-computed.}
    \label{fig:failure-regions}
\end{figure}

For the nonholonomic robot data, we construct $\mathcal{X}_1$ as a unit cube such that $x \in [-1.0, -2.0 ]$, $y \in [-2.25, -3.25]$, $\theta \in [1.25, 2.25]$, and $\mathcal{X}_2$ as a unit cube such that $x \in [0.75, 1.75]$, $y \in [-3.25, -4.25]$, $\theta \in [-1.0, -2.0]$. 
\Cref{fig:failure-regions} visualizes these regions in target space (left) and latent space (right).

For the racecar experiments, a small subset of the training data is considered to build the representative failure regions. 
$\mathcal{X}_1$ is constructed from the datapoints with $x > 0.0$, $\psi > 2.75$ (samples with a large yaw angle) and $\mathcal{X}_2$ is constructed from the datapoints with $x > 1.5$, $\dot{\psi} < -2.25$ (samples with a low yaw rate). 
Likewise, a small subset of the training data is used to build $\mathcal{X}_1$ and $\mathcal{X}_2$ for the F-16 experiments. 
$\mathcal{X}_1$ is obtained from the datapoints with a pitch $\theta > 1.45$ (high pitch) and $\mathcal{X}_2$ is constructed from the datapoints with an altitude $< -2.45$ (low altitude). 
Note that each state variable was normalized to have zero mean and unit variance.
Code to reproduce the experimental results is available at \texttt{\url{https://github.com/sisl/LatentImportanceSampling}}.

\subsection{Results}
\Cref{tab:metrics} presents the experimental results across the three datasets.
We run $100$ trials for the robot and racecar datasets and $20$ trials for the F-16 dataset, recording the metric means and standard deviations.
The latent sampling methods consistently achieve the lowest average negative log-likelihood values, indicating that the samples generated in latent space and pushed forward through the flow transformation more closely match the learned target density than points sampled directly in target space. 
Likewise, the latent-space IS methods achieve higher density and coverage scores than the target-space IS methods. 
The higher density scores indicate that latent IS places more samples in regions with a higher density of actual failure events. 
The higher coverage scores show that latent IS does a better job of finding all failure modes and generating samples in regions of the simulator outcome space that contain failure events. 
Performing the cross-entropy method in latent space results in the smallest average number of samples until convergence across all datasets, resulting in the lowest computational cost.

%%%%%%%%%%%%%%%%%%%%%%%%%%%%%%%%%%%%%%%%%%%%%%%%%%%%%%%%%%%%%%%%%%%%%%%%%%%%%%%%
% Conclusion
%%%%%%%%%%%%%%%%%%%%%%%%%%%%%%%%%%%%%%%%%%%%%%%%%%%%%%%%%%%%%%%%%%%%%%%%%%%%%%%%
\section{Discussion and Future Work}
Importance sampling is a powerful method for computing the probability of rare events and validating autonomous systems \cite{corso2021survey}. 
In this work we present a technique to improve importance sampling by first transforming the data with a normalizing flow. 
The invertible flow transformation warps non-isotropic target densities into a latent density which is more isotropic and easier to explore. 
We also propose an intuitive cost function that is simple to evaluate, even after undergoing an arbitrarily complex transformation to latent space. 
We experimentally show that conducting IS in latent space results in failure samples that more closely match the true distribution of failure events. 
Samples generated in latent space can be easily recovered in target space via the deterministic and invertible flow mapping. 
Our latent IS methods outperform target IS methods on a range of simulated autonomous systems.

Future work will provide a theoretical analysis of the benefits of latent IS, investigating the impact of maximum likelihood training on the warped geometry of the latent failure regions. 
Alternative cost-function formulations will be explored; for example, maximum-volume inscribed ellipsoids could result in $P_F$ estimates that are less conservative.
Finally, the latent IS methods presented in this paper will be used to validate real-world, black-box autonomous systems with high-fidelity simulators.

%%%%%%%%%%%%%%%%%%%%%%%%%%%%%%%%%%%%%%%%%%%%%%%%%%%%%%%%%%%%%%%%%%%%%%%%%%%%%%%%
% Acknowledgments
%%%%%%%%%%%%%%%%%%%%%%%%%%%%%%%%%%%%%%%%%%%%%%%%%%%%%%%%%%%%%%%%%%%%%%%%%%%%%%%%
\section*{Acknowledgments}
Toyota Research Institute (TRI) provided funds to assist the authors with their research, but this article solely reflects the opinions and conclusions of its authors and not TRI or any other Toyota entity. 
The NASA University Leadership Initiative provided funds to assist the authors with their research, but this article solely reflects the opinions and conclusions of its authors and not any NASA entity.

\bibliography{aaai25}

\end{document}